\documentclass[sigplan, twocolumn]{acmart} 
\renewcommand\footnotetextcopyrightpermission[1]{} 
\usepackage{amsmath}
\usepackage{graphicx}
\usepackage[frozencache,cachedir=minted-cache]{minted}
\usepackage{enumitem}

\begin{document}

\title{Distributed Graph Neural Network Inference With Just-In-Time Compilation For Industry-Scale Graphs}
\author{Xiabao Wu, Yongchao Liu, Wei Qin, Chuntao Hong}
\affiliation{%
  \institution{Ant Group, China}
  \city{}
  \country{}}
\email{{wuxiabao.wxb, yongchao.ly, johnny.qw, chuntao.hct}@antgroup.com}
\maketitle
Graph neural networks (GNNs) have delivered remarkable results in various fields. However, the rapid increase in the scale of graph data has introduced significant performance bottlenecks for GNN inference. Both computational complexity and memory usage have risen dramatically, with memory becoming a critical limitation. Although graph sampling-based subgraph learning methods can help mitigate computational and memory demands, they come with drawbacks such as information loss and high redundant computation among subgraphs.
This paper introduces an innovative processing paradgim for distributed graph learning that abstracts GNNs with a new set of programming interfaces and leverages Just-In-Time (JIT) compilation technology to its full potential. This paradigm enables GNNs to highly exploit the computational resources of distributed clusters by eliminating the drawbacks of subgraph learning methods, leading to a more efficient inference process.
Our experimental results demonstrate that on industry-scale graphs of up to \textbf{500 million nodes and 22.4 billion edges}, our method can produce a performance boost of up to \textbf{27.4 times}.

\section{Introduction}
As the graph data scale increases exponentially, GNN inference is encountering significant performance bottlenecks. The vast number of nodes and edges in large-scale graphs substantially amplifies core computational tasks such as graph convolution, message passing, and aggregation, leading to a notable decrease in inference speed. In addition, handling large-scale graph data requires substantial memory. Surpassing memory limits can result in memory overflow, compromising system stability and availability.

To mitigate these issues, practitioners often utilize graph sampling-based subgraph learning techniques~\cite{zhang2020agl}, which involve selecting specific nodes and edges to create mini-batches of subgraphs for inference, thereby curtailing computational and memory demands.
However, it has some notable drawbacks, such as causing information loss, decreasing inference accuracy, and introducing redundant computational overhead among subgraphs.

Moreover, JIT (Just-In-Time) compilation technology offers distinct advantages for enhancing inference acceleration and deployment. It compiles code into machine code during runtime, merging the benefits of interpretive execution with those of static compilation. In deep learning, JIT has enhanced the training and inference performance of conventional neural networks by operation fusion, storage and overhead reduction, and code optimization, thereby providing novel solutions for graph learning inference.

In this study, we introduce a novel distributed processing paradigm along with a new set of programming interfaces for GNN inference that overcomes the limitations of conventional subgraph learning methods. Our paradigm eliminates the need for subgraph extraction operations during inference on large-scale graph learning models, theoretically enabling the handling of graph data of any size. The key concept involves breaking down complex GNN models into multiple simple deep learning modules that function independently without requiring machine communication. By harnessing the full potential of JIT compilation technology, we can produce highly efficient executable programs, facilitating easy deployment and optimized performance.

\section{Methods}
Our paradigm is implemented based on DFOGraph~\cite{Yu2021DFOGraphAI} (integrated into our graph intelligent computing system~\cite{liu2023graphtheta} deployed in production), but introduces moderate modifications allowing for acquiring batches of dense data during processing, particularly in message passing.
We introduce a new set of programming interfaces, including two processing interfaces and two data retrieval functions for node and edge features.
In theory, our paradigm can be incorporated into any distributed graph processing engine.

Processing interfaces are \textbf{transform} and \textbf{message\_passing}, while data retrieval functions include \textbf{get$\_$vertex} and \textbf{get$\_$edge}.
\textbf{transform} is used mainly to process node or edge data locally, while \textbf{message\_passing} deals with the logic involved in remote message passing.
Notably, unlike existing frameworks such as PyG and DGL, users do not need to concern themselves with how messages are transmitted through edges during actual implementation, and also do not need to manually write or invoke functions related to message passing concerning edges. Instead, they are automatically managed by the runtime.
We implement our JIT compilation with \texttt{torch.jit}.
Upon decomposition, the graph algorithms convert sparse graph data into dense matrices during data processing, thus further improving inference performance.

\textbf{Programming Interfaces and JIT Compilation}
To clarify our design rationale, we have chosen some specific steps from the \texttt{GATConv} algorithm~\cite{velikovic2017graph} in PyG~\cite{Fey/Lenssen/2019} as an example. The following code snippet shows how the module performs linear projection on node features and aggregates them into new features based on edges.
We will elaborate the programming interfaces and methodologies below.
\begin{minted}[xleftmargin=20pt, fontsize=\tiny, linenos]{python}
import torch
from torch.nn.linear import Linear

class GATConvVertex(torch.nn.Module):
    def __init__(...):
      self.lin = Linear(...)
      self.att_src = Parameter(torch.empty(...))
      self.att_dst = Parameter(torch.empty(...))
      ...
    def forward(self, x):
      ...
      x_src = x_dst = self.lin(x)
      alpha_src = (x_src * self.att_src).sum(-1)
      alpha_dst = (x_dst * self.att_dst).sum(-1)
      return alpha_src, alpha_dst

class GATConvMP(torch.nn.Module):
    def __init__(...):
        ...
    def forward(self, src: Tensor, dst: Tensor):
        alpha = dst + src
        alpha = F.leaky_relu(alpha)
        return alpha
        
vertexmodel = GATConvVertex(...)
mpmodel = GATConvMP(...)
x = get_vertex("input")
alpha_src, alpha_dst = transform(x, vertexmodel)
alpha = message_passing(alpha_src, alpha_dst, mpmodel)
\end{minted}
\begin{itemize}[leftmargin=*]
\item \textbf{get\_vertex / get\_edge}: enables users to access the relevant data for processing based on the feature name. This interface merely returns a reference to the data and does not carry out any actual read or write operations (line 27).

\item \textbf{transform}: is primarily utilized for independently processing the features of individual nodes or edges. This interface does not require message passing and can be executed locally on each distributed compute node. It takes as input the features of the nodes or edges that are to be processed and the associated processing logic (line 28).

\item \textbf{message\_passing}: With this interface, users don't have to consider specific aggregation methods like \texttt{concat}, \texttt{add}, or others.
Its message passing is implicitly handled through the GAS model~\cite{Yu2021DFOGraphAI}, operating transparently so that users remain completely unaware of the underlying process.
They only need to use node or edge data flexibly according to their own needs just like handling dense tensors within a single-machine setup.
Line 29 demonstrates this, where the implemented functions only require simple addition, multiplication, or other operations such as \texttt{ReLU}.

\item \textbf{JIT compilation}: After the reconstruction of the PyG algorithm, it is transformed into the deep learning modules \texttt{GATConvVertex} and \texttt{GATConvMP}, optimized for independent dense tensor processing, as shown in the above code snippet. By utilizing torch.jit, efficient machine code can be produced. Depending on the application requirements, the code can be generated either offline or online.
\end{itemize}

\section{Results}
We utilize graphs from 3 real-world business scenarios in Ant Group and compare our method (on CPU docker clusters) with in-house graph sampling-based subgraph learning implementations (on V100 GPU clusters) of Ant Group. Our comprehensive comparison yields the following results: \textcircled{1} For the fraud detection graph, which consists of \textbf{340 million nodes and 1.1 billion edges}, the inference speed of the HGT model is improved by \textbf{12.8 times}. \textcircled{2} In the digital technology business graph, featuring \textbf{800 million nodes and 7.4 billion edges}, the GeniePath model gets its inference speed improved by \textbf{a factor of 8}. \textcircled{3} Regarding the credit graph, which includes \textbf{500 million nodes and 22.4 billion edges}, the performance of the GAT model shows a significant improvement of \textbf{27.4 times}.

The experimental results indicate that the improvements primarily arise from two key factors. Firstly, full-graph inference effectively mitigates the issue of both redundant computations and significant time consumption associated with graph sampling-based methods, especially for large-scale graphs.
Secondly, performance enhancements are achieved via JIT technology, which converts the original sparse graph data computation with communication involved into local dense data computation, and optimizes it into machine code.

\section{Conclusion}
The paper introduces a novel, efficient processing paradigm that distinctly organizes the workflow of GNNs, allowing each section to operate concisely.
This separation minimizes unnecessary communication, thereby simplifying and enhancing the processing efficiency.
By utilizing \texttt{torch.jit}'s robust JIT compilation capability, the solution generates highly efficient machine code. This optimization greatly improves the computational efficiency and reduces the inference time for GNNs, making deployment more friendly. Consequently, our paradigm simplifies complex GNN inference tasks into a more manageable and deployable process, offering substantial support for the advancement of industry-scale graph data processing and related applications.
Finally, it is worth mentioning that our paradigm has been deployed in production for more than two years.

\bibliographystyle{ACM-Reference-Format}
\bibliography{ref}


\begin{thebibliography}{5}


\ifx \showCODEN    \undefined \def \showCODEN     #1{\unskip}     \fi
\ifx \showISBNx    \undefined \def \showISBNx     #1{\unskip}     \fi
\ifx \showISBNxiii \undefined \def \showISBNxiii  #1{\unskip}     \fi
\ifx \showISSN     \undefined \def \showISSN      #1{\unskip}     \fi
\ifx \showLCCN     \undefined \def \showLCCN      #1{\unskip}     \fi
\ifx \shownote     \undefined \def \shownote      #1{#1}          \fi
\ifx \showarticletitle \undefined \def \showarticletitle #1{#1}   \fi
\ifx \showURL      \undefined \def \showURL       {\relax}        \fi
\providecommand\bibfield[2]{#2}
\providecommand\bibinfo[2]{#2}
\providecommand\natexlab[1]{#1}
\providecommand\showeprint[2][]{arXiv:#2}

\bibitem[Fey and Lenssen(2019)]%
        {Fey/Lenssen/2019}
\bibfield{author}{\bibinfo{person}{Matthias Fey} {and} \bibinfo{person}{Jan~E.
  Lenssen}.} \bibinfo{year}{2019}\natexlab{}.
\newblock \showarticletitle{Fast Graph Representation Learning with {PyTorch
  Geometric}}. In \bibinfo{booktitle}{\emph{ICLR Workshop}}.
\newblock


\bibitem[Liu et~al\mbox{.}(2023)]%
        {liu2023graphtheta}
\bibfield{author}{\bibinfo{person}{Yongchao Liu}, \bibinfo{person}{Houyi Li},
  \bibinfo{person}{Guowei Zhang}, \bibinfo{person}{Xintan Zeng},
  \bibinfo{person}{Yongyong Li}, \bibinfo{person}{Bin Huang},
  \bibinfo{person}{Peng Zhang}, \bibinfo{person}{Zhao Li},
  \bibinfo{person}{Xiaowei Zhu}, \bibinfo{person}{Changhua He}, {and}
  \bibinfo{person}{Wenguang Chen}.} \bibinfo{year}{2023}\natexlab{}.
\newblock \bibinfo{title}{GraphTheta: A Distributed Graph Neural Network
  Learning System With Flexible Training Strategy}.
\newblock
\showeprint[arxiv]{2104.10569}


\bibitem[Veličković et~al\mbox{.}(2018)]%
        {velikovic2017graph}
\bibfield{author}{\bibinfo{person}{Petar Veličković},
  \bibinfo{person}{Guillem Cucurull}, \bibinfo{person}{Arantxa Casanova},
  \bibinfo{person}{Adriana Romero}, \bibinfo{person}{Pietro Liò}, {and}
  \bibinfo{person}{Yoshua Bengio}.} \bibinfo{year}{2018}\natexlab{}.
\newblock \showarticletitle{Graph Attention Networks}. In
  \bibinfo{booktitle}{\emph{ICLR}}.
\newblock


\bibitem[Yu et~al\mbox{.}(2021)]%
        {Yu2021DFOGraphAI}
\bibfield{author}{\bibinfo{person}{Jiping Yu}, \bibinfo{person}{Wei Qin},
  \bibinfo{person}{Xiaowei Zhu}, \bibinfo{person}{Zhenbo Sun},
  \bibinfo{person}{Jianqiang Huang}, \bibinfo{person}{Xiaohan Li}, {and}
  \bibinfo{person}{Wenguang Chen}.} \bibinfo{year}{2021}\natexlab{}.
\newblock \showarticletitle{DFOGraph: an I/O- and communication-efficient
  system for distributed fully-out-of-core graph processing}. In
  \bibinfo{booktitle}{\emph{PPoPP}}.
\newblock


\bibitem[Zhang et~al\mbox{.}(2020)]%
        {zhang2020agl}
\bibfield{author}{\bibinfo{person}{Dalong Zhang}, \bibinfo{person}{Xin Huang},
  \bibinfo{person}{Ziqi Liu}, \bibinfo{person}{Jun Zhou},
  \bibinfo{person}{Zhiyang Hu}, \bibinfo{person}{Xianzheng Song},
  \bibinfo{person}{Zhibang Ge}, \bibinfo{person}{Lin Wang},
  \bibinfo{person}{Zhiqiang Zhang}, {and} \bibinfo{person}{Yuan Qi}.}
  \bibinfo{year}{2020}\natexlab{}.
\newblock \showarticletitle{AGL: A Scalable System for Industrial-purpose Graph
  Machine Learning}.
\newblock \bibinfo{journal}{\emph{Proceedings of the VLDB Endowment}}
  \bibinfo{volume}{13}, \bibinfo{number}{12} (\bibinfo{year}{2020}).
\newblock


\end{thebibliography}

\end{document}